# Neural Approaches for Data Driven Dependency Parsing in Sanskrit

**Amrith Krishna, Ashim Gupta, Deepak Garasangi,**
**Jivnesh Sandhan, Pavankumar Satuluri, Pawan Goyal**
ITU Copenhangen, University of Utah, Manipal Institute of Technology,
IIT Kanpur, Chinmaya Vishwavidyapeeth, IIT Kharagpur
`amrk@itu.dk, pawang@cse.iitkgp.ac.in`

## Abstract

Data-driven approaches for dependency parsing have been of great interest in Natural Language Processing for the past couple of decades. However, Sanskrit still lacks a robust purely data-driven dependency parser, probably with an exception to Krishna (2019). This can primarily be attributed to the lack of availability of task-specific labelled data and the morphologically rich nature of the language. In this work, we evaluate four different data-driven machine learning models, originally proposed for different languages, and compare their performances on Sanskrit data. We experiment with 2 graph based and 2 transition based parsers. We compare the performance of each of the models in a low-resource setting, with 1,500 sentences for training. Further, since our focus is on the learning power of each of the models, we do not incorporate any Sanskrit specific features explicitly into the models, and rather use the default settings in each of the paper for obtaining the feature functions. In this work, we analyse the performance of the parsers using both an in-domain and an out-of-domain test dataset. We also investigate the impact of word ordering in which the sentences are provided as input to these systems, by parsing verses and their corresponding prose order (anvaya) sentences.

## 1 Introduction

Dependency parsing, specifically data-driven dependency parsing, has garnered considerable attention from the computational linguistics and the Natural Language Processing (NLP) communities in the past few decades. The ability of the parsing approach to capture syntactic and shallow semantic information over multiple languages and often in cross-lingual settings has led to widespread adoption of the syntactic parsing approach (Tiedemann, 2015; Li et al., 2019; McDonald et al., 2013). Transition based parsing and Graph based parsing are two of the dominant approaches in data-driven dependency parsing (McDonald and Nivre, 2007). Parsers proposed under either of these approaches vary widely in terms of the input representations, learning paradigms and inferences employed. The parsers range from linear machine learning models to neural network based nonlinear models for learning, whereas the input representations vary from discrete hand-crafted features to continuous word representations such as pre-trained word embeddings, contextualised word representations etc. (Kulmizev et al., 2019; Peters et al., 2018). In this work, we evaluate and present a comparative error analysis of the performance of several of these recently proposed data-driven approaches for parsing sentences in Sanskrit.

Our work closely follows the analysis framework proposed in several of the previous works on dependency parsing. McDonald and Nivre (2007), which was later extended in McDonald and Nivre (2011), performed an empirical analysis which contrasts the performance of graph-based and transition based parsers. They identified characteristics of these parsers and profiled the structural divergences in the predictions and the errors made by each of them. This was further extended by Kulmizev et al. (2019), where they performed a similar analysis on neural dependency parsers. Further, they investigated the impact of using contextualised word representations (Peters et al., 2018). Here, they observed that with neural architectures and contextualised word embeddings, both graph based and transition based approaches, tend to converge on their performance characteristics more as compared to the corresponding linear models. The analysis framework essentially measures the empirical performance of various models based on the sentence length, relative distance between words connected by a dependency arc, distance of nodes to the root and on non-projective arcs. In this work, we replicate these analyses exclusively for Sanskrit. Further, we add an additional dimension crucial for Sanskrit, i.e., analysis of differences in parsing poetry and prose data.

| Parser | Parsing Approach | Learning Framework | Input Representation |
|---|---|---|---|
| YAP (More et al., 2019) | Transition | Neural | Hand-crafted features |
| Dozat and Manning (2017) | Graph | Neural | Word representations |
| Self-training (Rotman and Reichart, 2019) | Graph | Neural | Contextualised word representations |
| Learning to Search (Chang et al., 2016) | Transition | Neural | Hand-crafted features |

Table 1: Overview of the data-driven models evaluated in this work.

Table 1 gives an overview of the various systems we use along with the parsing approach, model and input representation used in these models. Here, all the four models use neural learners. However, both the transition based models we evaluate use hand-crafted features while the graph based parsers automatically learn the word representations as features. Neural parsers, both graph-based and transition based, that use the learned word representations as features currently report the state of the art results for many languages including English (Mrini et al., 2019; Qi et al., 2018).[1] However this need not hold true for morphologically rich and relatively free word order languages. For instance, YAP (More et al., 2019) currently reports state of the art parsing results for the morphologically rich language, Hebrew and hence we include the model in our evaluation. YAP makes use of word n-grams and morphological information of the words from the input sentence as features. Similarly, we chose to include the Learning to Search model due to its training procedure that is designed to mitigate the effects of exposure bias and label bias. Since the prediction of individual arcs are dependent on the previous arc predictions for a sentence in dependency parsing, Learning to Search (Chang et al., 2016) exposes the system to incorrect parsing decisions during training. This enables the system to recover from early incorrect predictions made during inference and thereby resulting in lower exposure bias and label bias (Wiseman and Rush, 2016). The graph based parsers we employ learn neural word representations as features. This results in requirement for large amount of training data, or availability of large monolingual corpora to obtain pre-trained embeddings (Che et al., 2018). Rotman and Reichart (2019) addresses this issue by proposing a self-training approach that augments the biaffine parser from Dozat and Manning (2017). Here, first a base parser is trained with as low as 500 trained samples and then, dependency trees are predicted for unlabelled data. The predictions are used to train on auxiliary tasks which results in contextualised representations of the input sequence based on those auxiliary tasks. As Rotman and Reichart (2019) is promising as a parser for low-resource languages, we include the model in our experiments. Further, since Dozat and Manning (2017) is the base parser to Rotman and Reichart (2019), we include the model as well for a fair comparison between the models. Based on our experiments, our major observations are:

1. The self-training based dependency parser by Rotman and Reichart (2019) performs the best when tested on our in-domain test dataset (Kulkarni, 2013) and also on the out of domain Śiśupāla-vadha dataset, with word arrangements in prose as well as poetry.

---
[1] https://nlpprogress.com/english/dependency_parsing.html

2. All the models provide competitive results even on out-of-domain Śiśupāla-vadha test data, when the sentence follows prose ordering. However, the performance of all the models falls drastically, often halves, when the same sentences from Śiśupāla-vadha in its verse ordering are provided as input during inference.

3. We find that use of automated feature functions learned using LSTM based word-representation can be benefited in a low-resource morphologically rich language like Sanskrit. Both such graph based parsing models we use (Dozat and Manning, 2017; Rotman and Reichart, 2019) are the best 2 models in our experiments. Transition based parsers in general tend to perform poorly when it comes to predictions on longer sentences. However, the learning to search transition parser remains relatively stable often competitive to the graph based parsers we use.

## 2 Data-Driven Dependency Parsing

Early attempts at data-driven models for parsing dependency structures could be traced back to the later half of the 1990s. Collins (1996) and Eisner (1996), proposed multiple probabilistic-models for dependency parsing of English sentences. Similarly, Kudo and Matsumoto (2000) proposed an SVM based classifier for classifying dependencies in Japanese, and this was later extended to English (Yamada and Matsumoto, 2003). In the tenth CoNLL-X shard task on multi-lingual dependency parsing (Buchholz and Marsi, 2006), it was observed that graph-based and transition based parsing approaches have emerged to be the two dominant approaches for parsing dependency structures. Though the shared task identified these approaches as "all-pairs" and "stepwise", McDonald and Nivre (2007) later termed these approaches as graph based and transition based. Even though most of the proposed parsers fall under either of these categories, these parsers vary widely in terms of the input representations, learning algorithms and learning architectures. In this section, we provide a brief overview of the graph based and transition based parsers along with the developments that occurred in the past couple of decades. For a detailed understanding of these approaches, we recommend the readers to refer to McDonald and Nivre (2011) and Kulmizev et al. (2019).

### 2.1 Graph-Based Parsing

Given an input sentence, the sentence is converted into a dense graph (or a complete graph), where each word in the sentence is a node and every possible dependency relation between two words forms a labelled arc. In this 'all-pairs' approach (Buchholz and Marsi, 2006) every possible word pair is scored based on the likelihood of the pair to form a dependency relation. Then, using a search mechanism the parser identifies the best parse for the given sentence. We can generalise graph-based dependency parsing as an approach to score a dependency tree by the linear combination of scores of its local sub-graphs. The task here is to find an induced subgraph of the input graph, which structurally is a tree that spans all the nodes of the graph, i.e. a spanning tree. The inference then translates to finding the spanning tree with maximum (or minimum) score. The score for a tree is obtained by decomposing the structure into sub-graphs, and is scored using a global optimisation approach with a non-local greedy search. The decomposed sub-graphs can be as simple as an edge that connects a pair of nodes, and such models are called arc-factored models. The learning involves creating a gap between the score of the ground truth tree with that of other candidate tress. As per Kübler et al. (2009), any graph-based parsing system must define four things:

1. A set of sub-structures $\Psi_G$, to which the input graph $G$ will be decomposed.

2. A defintion of the parameters $\lambda$ for each substructure in $\Psi_G$.

3. A method for learning $\lambda$.

4. A parsing algorithm.

In an arc-factored model, $\Psi_G$ will be the edge set in the input graph $G$ and $\lambda$ will be a function that maps edges into real values. Here an edge can be represented as a vector of fixed features and the function will be learnt to map this vector into a scalar real-value. The score of candidate structures (spanning trees in this case) will be calculated by summing the score of their edges. The parsing algorithm is then reduced to searching for a suitable structure, generally a spanning tree. The search procedure can be modified to enforce additional constraints on the specific structure to be searched. Additional constraints, such as the structure should only have one edge labelled with the subject relation, and relaxations, such as allowing cycles on the predicted structures, were incorporated in the inference in several works (Kübler et al., 2009). McDonald et al. (2005b) formalised the search procedure as a search for spanning tree with maximum score. Here the score of the tree is nothing but the sum of the scores of its edges. They employed the Chu-Liu-Edmond's algorithm (Edmonds, 1967).

Arc-factored models , such as McDonald et al. (2005b), perform exact search and hence they do not suffer from search errors. However, the features that can be incorporated are restricted to the local sub-graphs. McDonald et al. (2005a) used a feature function that contains 18 feature templates relevant to each of the edges. The features for an edge essentially are the word and POS information of the nodes connected by the edge, and the word and POS information of adjacent words to these nodes as per the linear order of the sentence. Although we learn a function to score the individual edges, the training still is global as the parameters of the model are set relative to the classification of the entire dependency tree, and not locally over single edges (McDonald and Nivre, 2011). The graph parsers are typically trained using margin based structured learning approaches which attempt to widen the gap between the score (need not always be a probabilistic score) of the correct dependency tree to that of other candidate dependency trees.

Graph-based parsers were extended to higher-order models and also in terms of the arity of a node. The arity of a node measures the likelihood of the node to have a fixed number of dependents. These extensions were primarily to improve the parsing performance for non-projective dependency structures. The arity information was trivially incorporated by modifying the scoring function of arc-factored models by adding the likelihood of arity as an additional parameter (Kübler et al., 2009). In the case of higher order-models, McDonald and Pereira (2006), score adjacent arcs with a common head. Similarly Carreras (2007) considered head-modifier chains of length two. Third order models by Koo and Collins (2010) consider set of various three arc subgraphs, such as adjacent siblings or grand-siblings. However, the inference and learning becomes NP-Hard (McDonald and Satta, 2007) when the score is factored over higher order structures than edges, especially for non-projective dependency parses. Martins et al. (2009) and Martins et al. (2010) proposed parsing approaches with exact inference, though with NP-hard complexity, to incorporate higher order features. To make the inference more efficient, an approximate inference using cube pruning was introduced in Zhang and McDonald (2014). However recent neural approaches reverted to using first order models and still achieve state of the art results (Kulmizev et al., 2019).

## 2.2 Transition based parsing

In transition based parsing approaches, instead of considering all the possible word pairs, the parsing decisions are taken incrementally. The approach is inspired from shift-reduce parsing approaches originally proposed for analysing programming languages (Aho and Ullman, 1972). Essentially this reduces the parsing task to that of iteratively predicting the next immediate parsing action as a greedy search. The system is trained on partial parser configurations, where the immediate next action needs to be predicted. Transition based parser parses a sentence sequentially from left to right.[2]

---

[2] The parsing can be done sequentially from right to left as well.

The parsing system is an abstract machine, consisting of a set of states and transitions between them. We start from an initial state and then follow a valid sequence of transitions, terminating at one of the terminal states. By the time a terminal state is reached, it would lead to the dependency tree for the given sentence. The parser essentially consists of a stack, a buffer and a set of dependency labels. At a given instance, the stack maintains the partial parses, the buffer maintains the words to be parsed and additionally the labels assigned so far is kept track of. The combination of the configuration of these three entities result in a parser configuration, i.e. a state. The parser is initialised with empty stack with a dummy root node in it, a buffer with all the words in the sentence and an empty set of actions. Words one after other are pushed into the stack and popped from it. A final configuration is a configuration when the buffer is empty and the stack is left only with the dummy root node. The given system is a non-deterministic system, i.e. there can be multiple possible transitions from a given state. However, the system can be amenable to a deterministic parsing approach, assuming the presence of an oracle that can guide the transition decisions at each state. The oracle can be approximated by training classifiers using labelled data (Kübler et al., 2009) and guide the parsing process.

Summarily, given a classifier, the parsing procedure is about finding a trajectory, i.e. an optimal sequence of transitions, that leads from the initial configuration to a terminal configuration. The behaviour of the parser is defined in terms of various transitions (McDonald and Nivre, 2011). Apart from initialisation and termination transitions, the basic version of the parser included four more transitions. The other transitions are Shift, reduce, left-arc, right-arc. The shift operation simply pushes the next input word into the stack, while reduce pops the word from the top of the stack. The left-arc assigns an arc from the next input word to word on the top of the stack and right-arc performs vice versa (McDonald and Nivre, 2007). Transition parsers enable to incorporate non-local features into the configurations of the parser. However, since the approach performs a local search, it is exposed to label bias resulting in compounding of errors. Though initially, transition based parsers employed greedy search approaches, models with global learning were later proposed. Further McDonald and Nivre (2011) proposed an enriched feature function with 72 different feature templates. The features vary from uni-gram to tri-gram word and POS information of words in the stack and the buffer, distance between the word at the top of the stack and front of the buffer queue, valency, third order features and set of dependency labels with the words at the top of the stack and the front of the buffer queue. More et al. (2019) adapted the feature function to meet the needs of free-word order languages, by replacing positional information with labelled grammatical functions.

In Goldberg and Nivre (2013), the static oracle was replaced with dynamic oracles. Dynamic oracles could guide the system to optimal trajectory, even from a configuration which is not present in the ground-truth. Introduction of dynamic oracles could reduce the error propagation, especially when an early mistake in prediction is made. While the transition based approaches initially were trained to handle projective dependency structures, these were later extended to handle non-projective dependency structures. Further, modifications to transition approaches were proposed, primarily for English. Apart from Arc-standard, which we already defined, arc-eager and arc-(Z)eager are two other systems of transition based parsing. The arc-eager system allows a right-dependent to eagerly attach to the head. Here, the dependent word will be pushed onto the stack so as to enable attaching to its dependents. Arc-(Z)eager is an augmentation to arc-eager approach, where it employs an additional stack to hold the head nodes and enabling certain constraints for the head words during transitions (More et al., 2019).

## 2.3 Neural Dependency Parsing

Though neural network based approaches for dependency parsing have been explored towards the end of the last decade (Titov and Henderson, 2007; Attardi et al., 2009), these parsers have come to prominence since the last five years (Kulmizev et al., 2019). Use of neural approaches for dependency parsing has brought in great advancements in the learning methods, architectures and improvements in performance. However, most of these parsers still belong to graph based or transition based approaches. Neural parsers have enabled to bring down the complexities in the parsing approaches and revert back to simpler inference approaches without any loss in performance. We can observe that neural transition parsers with local learning and greedy inference, and similarly neural graph based parsers making use of first order models now report competitive, often better, results as compared to complex parsing approaches previously employed. Yet another major advantage with neural architectures is that they have alleviated the need for hand-crafted feature engineering, and instead started to rely on dense features in the form of embeddings. Chen and Manning (2014) and Weiss et al. (2015) have shown how neural network based dense features can be employed instead of hand-crafted features for transition based and graph-based parsers respectively. Kiperwasser and Goldberg (2016) has employed BiLSTM based word features, coupled with POS tags, that could capture the contextual representation of the words within the sentence. However these models are constrained by the need for high amount of training data. Moreover, for morphologically rich languages like Hebrew or Turkish, state of the art models are the ones that still use hand-crafted features (More et al., 2019; Seeker and Çetinoğlu, 2015).

## 3 System Description of the Parsing Models

The Pāṇinian theory of kāraka, proposed for dependency analysis of sentences in Sanskrit, is well studied in the computational linguistics community. Bharati et al. (1991) proposed a computational framework for multiple Indian languages based on this framework. Further a dependency tagset and a dependency parser using the tagset was developed over the years for multiple Indian languages (Tandon and Sharma, 2017; Begum et al., 2008). Rule based systems for dependency parsing in Sanskrit was proposed by Kulkarni et al. (2010) and was later improved in Kulkarni (2013) using dynamic programming approaches. However both the systems require the input sequence to be in prose (anvaya) order, as the systems build constraints based on the (relative) positional information of the words in the sentence. A dependency parser for verses in Sanskrit was later proposed in Kulkarni et al. (2019). A statistical data-driven dependency parser was proposed by Krishna (2019), which uses an energy based model for the dependency parsing. All of these models use the tagset based on the relations described in Ramkrishnamacharyulu (2009). In this work, we do not consider any of the aforementioned works for comparison, as we focus on completely data-driven approaches with no explicit linguistic information incorporated into the models. Hence, we compare four different systems as shown in Table 1. While all of them use a neural network for training the models, 2 of them still use hand-crafted features and the other two rely on LSTM based encoders for capturing the feature information about the words and the sequence information of the words in context.

### 3.1 Yet Another Parser (YAP)

The parser proposed by More et al. (2019), is a transition based parser. It currently reports the state of the art results for Hebrew, a morphologically rich language. The model uses hand-crafted features, adapted from Zhang and Nivre (2011) for free word order languages. The model uses a variant of the generalised perceptron (Zhang and Clark, 2011) with early-update (Collins and Roark, 2004) for training. The focus of feature engineering is on selectional preferences, sub-categorisation frames and their distributional characterisation. The transition based system has provisions for Arc-standard, Arc-eager and Arc-(Z)eager transition paradigms. We use the Arc-eager apparoach in our models. While the paper originally proposes a joint morpho-syntactic parser that can jointly perform morphological parsing and syntactic parsing for Hebrew, we make use of the system as a standalone dependency parser.

### 3.2 Learning to Search (L2S)

Learning to search is a framework that enables the formulation of a structured prediction problem as a search problem (Ross et al., 2011; Collins and Roark, 2004; Daumé et al., 2009). Here, an explicit search space is generally defined as a finite state machine over a set of states. The search problem essentially consists of a set of states, including an initial state, terminal states and actions that enable transition from one state to another. The learning problem here is to learn a function, referred to as a policy, that maps a trajectory (sequence of actions) from the initial state to a goal state with the lowest cost or dually the highest score. Here each state configuration is represented using a bag of features, which is used for learning. L2S integrates the learning and prediction in a unified framework. This framework is similar to a reinforcement learning setting, and indeed in many ways it is. However, L2S uses a reference policy obtained from labelled examples before the training starts (Chang et al., 2015). During training, the supervision in the form of labelled data is used to define the loss function and to construct a reference policy (Chang et al., 2016). Based on these signals, the classifier (or the current policy) gets iteratively trained.

Once a search space is obtained for a given problem, the training can be started. It needs to be noted that the loss during training is not calculated based on individual actions, but based on the trajectory of actions taken for a given data point. In our case, the data point is a sentence, For a sentence, the training happens in 3 phases called as roll-in, roll-out and one-step deviation. The roll-in and roll-out policies can be the reference policy, current policy or a mixture of both. Since, the prediction at a given step is dependent on the history of predictions made so far for the input sentence, each of those mistakes has a different cost to pay. This is called a credit assignment problem. To handle such cases and thereby alleviate the exposure bias of such systems, we do a one step deviation at several all time steps during different iterations. The action taken during the deviation would be an alternate action from the set of possible actions. All the actions prior to the deviation follow from the roll-in policy and the actions after the deviation follows the roll-out policy. The loss is calculated based on the trajectory obtained as a consequence of these 3 steps.

We employ a transition based parsing approach for the L2S framework. Here, the state space consists of all possible configurations of the parser. As with other standard transition based parsers (Kuhlmann et al., 2011), we use the word-level and POS-level (morphological tags) information of upto three words in the stack and the buffer to construct the feature representation for each state. We follow the arc-hybrid system proposed by Goldberg and Nivre (2013) and the corresponding L2S implementation from Chang et al. (2016). The optimal reference policy is computed using the dynamic oracle approach by Goldberg and Nivre (2013). The loss is computed by computing the differences in the number of parents in the predicted tree and that of the gold-standard tree. Here, when an action is predicted, the dependency relation is also predicted whenever the action is a reduce operation.

### 3.3 BiAFFINE Parser (BiAFF)

The BiAFFINE parser (Dozat and Manning, 2017) is a neural graph-based dependency parsing approach similar to Kiperwasser and Goldberg (2016). The neural approaches, in general, have been beneficial for dependency parsing as it considerably reduces the complexities involved in generating feature representations. The feature representations are learned as embeddings from the surface forms of the words and also from their POS. For learning contextual embeddings that can capture the sentence context, LSTM (or BiLSTM) layers are used. Here, the authors introduce bi-affine attention instead of bilinear or MLP based attention, which were used in previous works in dependency parsing. The representations learned from LSTMs are subjected to MLP based dimensionality reduction and a biaffine dependency label classifier is used. Here, a biaffine classifier differs from an affine classifier in the following way. For an input vector $\vec{x}$, while an affine classifier could be expressed as $W\vec{x} + b$, then a biaffine classifier could be expressed as $W'(W\vec{x} + b) + b'$.

The model inputs a sequence of tokens and their POS tags, which is then passed through multiple layers of a BiLSTM network. In order to obtain specialised representations, the hidden vector from the final layer of the LSTM network is passed to four separate ReLU activation layers. This produces four separate representations, each for a separate purpose. One for word as a dependant looking for its head and another for its label. Similarly, we have one each for a word as head looking for its dependents and the labels of those dependents (Dozat and Manning, 2017). Two biaffine classifiers are used, where these representations are passed. The classifiers are used to predict the head and the label, respectively for a given word pair.

### 3.4 Deep Contextualized Self-training (DCST)

Deep Contextualized Self-training (Rotman and Reichart, 2019) extends the graph-based parser of Dozat and Manning (2017) with a self-training approach (Clark et al., 2018; Rybak and Wróblewska, 2018) for low resource dependency parsing. The work primarily addresses the scarcity of task-specific labelled data for low resource languages (or scarcity of in-domain data) and suggests a self-training based solution. Here, a base-learner is trained on a limited set of labelled data and then applied on a larger set of unlabelled data. Then the predictions from unlabelled data and the gold-standard labelled data are combined to retrain the final model, which is often performed iteratively. Specifically, in this work, it first trains the BiAFFINE parser as a base learner, with as low as 500 labelled training sentences and then uses the same to generate predictions from unlabelled data. The predicted dependency trees from the unlabelled data are then used to train neural networks for several auxiliary tasks. The auxiliary tasks performed are sequence level tagging tasks, where the input is the predicted dependency tree adapted to a suitable sequence tagging scheme. LSTM based taggers are used for training these tasks. Once the training is completed, the BiLSTM encoder layers are used for generating representations for the input word. The encoder layers from the auxillary tasks are combined with the representations from the encoder layers of the BiAFFINE parser using a gating mechanism (Sato et al., 2017). Here, during training, the representation layers of the parser are randomly initialised, while the representation layers from the auxiliary task are initialised to what it learned for the auxiliary task. The auxiliary tasks that are employed are, predicting the number of children for each word, predicting the distance of the word from the root in the tree and predicting the head word of a given word using the relative POS tagging scheme (Rotman and Reichart, 2019). Here, the relative POS tagging scheme enables to mark ahead word by its relative position from the current dependent node and the corresponding POS tag of the head word.

The basic intuition behind the approach in this work is similar to pretraining and multi-task approaches used in deep learning networks previously. However, in pretraining settings, though unlabelled data are used, they are used to train the LSTM encoders on language model based objective. In this work, they are trained on sequence level tagging tasks based objectives. This is similar to multi-task learning approaches where several related tasks are trained together or are reused, but those approaches typically use labelled data, unlike in this case.

## 4 Experiments

### 4.1 Dataset

The dataset for the dependency parsing is obtained from the department of Sanskrit studies, UoHyd[3]. We use about 1,500 prose sentences from the Sanskrit Tree Bank Corpus, henceforth to be referred to as STBC (Kulkarni et al., 2010; Kulkarni, 2013). Further, we use 1,000 sentences from STBC as the test data. Additionally, we take 300 sentences from the Śiśupāla-vadha corpus[4] for testing. Here, we make use of the prose order (anvaya) and the corresponding verse order of the sentences to provide as input to the different models. For all the models, we provide the segmented surface-form of the words in the sentence and their corresponding morphological tags as input. However, the entries in STBC corpus did not contain morphological tags for the words. For these sentences, we used the Sanskrit Heritage reader (Goyal and Huet, 2016; Huet and Goyal, 2013), a shallow parser, for obtaining the exhaustive enumeration of all the possible morphological analyses of the input. We only selected those sentences where none of the words expressed syncretism or homonymy. While, the current morphological tag information is yet to be verified by a domain expert, we decided to include the morphological tag information in input. This decision was motivated by the empirical observation that the inclusion of morphological tags as input results in about 3 to 4 % of performance improvement for all the models, as compared to using only the surface-forms of the words as input. Moreover, the evaluation of the models is based only on the dependency arcs and the labels, and hence this decision does not interfere in our experimental settings.

---

[3] http://sanskrit.uohyd.ac.in/scl/
[4] http://sanskrit.uohyd.ac.in/scl/e-readers/shishu/

|   | System | STBC Prose | | Śiśupāla-vadha Prose | | Śiśupāla-vadha Poetry | |
|---|--------|------|------|------|------|------|------|
|   |        | UAS  | LAS  | UAS  | LAS  | UAS  | LAS  |
| 1 | L2S    | 77.29 | 68.53 | 75.42 | 68.48 | 38.80 | 31.76 |
| 2 | YAP    | 72.14 | 63.15 | 65.27 | 55.88 | 26.79 | 21.10 |
| 3 | BiAFF  | 78.97 | 70.83 | 76.77 | 70.24 | 38.45 | 34.32 |
| 4 | DCST   | **80.97** | **72.86** | **81.08** | **74.37** | **40.02** | **35.70** |

Table 2: Performance evaluation of the four dependency parsers.

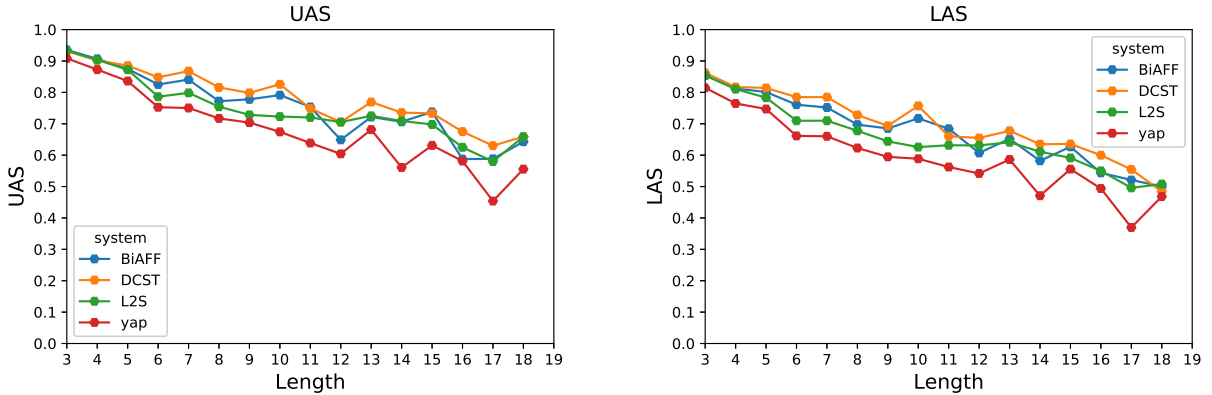

(a) Sentence length UAS - STBC  (b) Sentence length LAS - STBC

Figure 1: UAS/LAS of sentences grouped based on their sentence length on STBC corpus

### 4.2 Results

We report the performance of each system using the micro averaged Unlabelled Attachment Score (UAS) and the Labelled Attachment Score (LAS) as metrics (Nivre and Fang, 2017). UAS is the fraction of correct tokens with correct predicted heads and LAS is the fraction of tokens with the correct label (McDonald and Nivre, 2011). For micro averaged UAS and LAS, the denominator for normalising these fraction scores is the number of tokens in the corpus. Table 2 shows the results for all the four systems tested on the prose sentences from STBC, and prose and poetry data from Śiśupāla-vadha. It can be observed that DCST outperforms all the other models in all the three test data settings. The DCST model uses BiAFF as a base learner. However, it improves the performance of the model, by using better input representations obtained by self-training. We find that the performance differences between both the models are statistically significant.[5] Since, all the models are trained on the STBC train corpus, the STBC test data should ideally be most similar in terms of the data distribution with that of the training data. The data from Śiśupāla-vadha corpus can be seen as an out of domain dataset. Sentences from Śiśupāla-vadha, when arranged in their prose order, report comparable results with that of the STBC prose test data for all the models, with YAP being a possible exception. However, the performance for all the models drops drastically when these systems are provided with poetry order data as input. In fact, for all the systems, the performance drops to half the UAS as compared to the corresponding prose order input. This is a strong indication that the order in which the data is input affects the performance of these systems.

---
[5]We performed t-test for statistical significance with $p < 0.01$.

**Comparison to Linguistically involved Energy Based Model:** We compare the performance of all the four models with the performance of the linguistically involved energy based model (EBM) proposed by Krishna (2019). We report two results based on the EBM model. Both the models use linguistic information in its input representation and in its feature engineering. However EBM* uses linguistic information to prune the search space and filter the candidates during the inference. Both the models use a graph-based input and are agnostic to the order in which a sequence is provided. Further, they report the

|   | System | UAS | LAS |
|---|--------|------|------|
| 1 | L2S    | 81.9 | 72.46 |
| 2 | YAP    | 75.3 | 66 |
| 3 | BIAFF  | 82.35 | 75.65 |
| 4 | DCST   | 84.36 | 76.8 |
| 5 | EBM    | 82.65 | 79.28 |
| 6 | **EBM*** | **85.32** | **83.93** |

Table 3: Macro averaged UAS and LAS for 1300 sentences from the STBC and Śiśupāla-vadha prose test datasets.

Macro averaged UAS and LAS, i.e. average of sentence level UAS and LAS of sentences from both the STBC and Śiśupāla-vadha test data. Table 3 provides the macro averaged UAS and LAS for all the six systems. Here, we use the prose ordered data from Śiśupāla-vadha and STBC for the four models discussed in this paper. From the table, it can be observed that the EBM* model outperforms all the other models and reports a huge improvement in terms of LAS. However, DCST still outperforms the EBM, the model that does not incorporate explicit linguistic information during inference, in terms of UAS. This not only shows the relevance of incorporating linguistic information into the model, but also stresses on the significance of using contextualised word representations (like in the case of DCST).

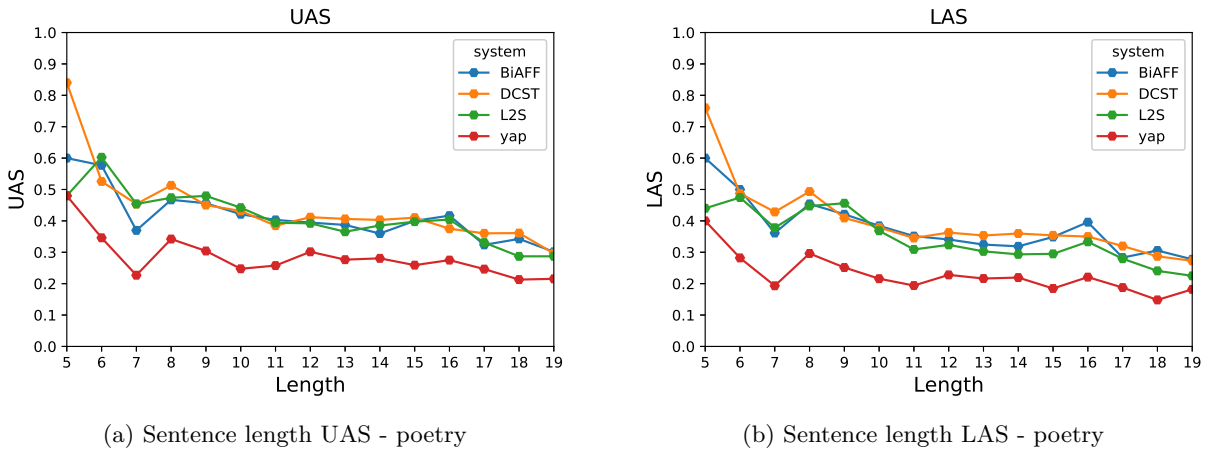

(a) Sentence length UAS - poetry

(b) Sentence length LAS - poetry

Figure 2: UAS/LAS of sentences in poetry order, grouped based on their sentence length on Śiśupāla-vadha corpus

**Sentence Length** Figure 1(a-b) and Figure 2(a-b) show the UAS and LAS for the sentences grouped based on the number of words in them, for each dataset. We show the lengthwise performance of only the poetry ordered sentences for the Śiśupāla-vadha dataset, as the prose ordered data's performance is similar to that of STBC prose test data. The figures show the performance for only those sentence lengths for which there exist at least 5 sentences in the corpus. All the systems perform the best on shorter sentences and generally the performance drops as the length of the sentence increases. However, the L2S system reports the lowest Mean

Absolute Deviation[6] (MAD) among the 4 models (0.07), followed by DCST (0.077) for the STBC corpus. YAP performs the worst and has the highest MAD (0.093). In the case of poetry data, the performance of the DCST dips drastically for sentences longer than 5 words as compared to its performance on sentences with 5 or less words. It shows the highest MAD (0.122) as well. The L2S system reports the lowest MAD among all the models (0.081). All the other systems have a MAD < 0.085. Sentences with 3 to 7 words form 68.7 % of sentences in the STBC corpus. In the case of Śiśupāla-vadha, 50 % of the sentences consist 11 to 15 words, and 85 % of the sentences consist of 7 to 17 words.

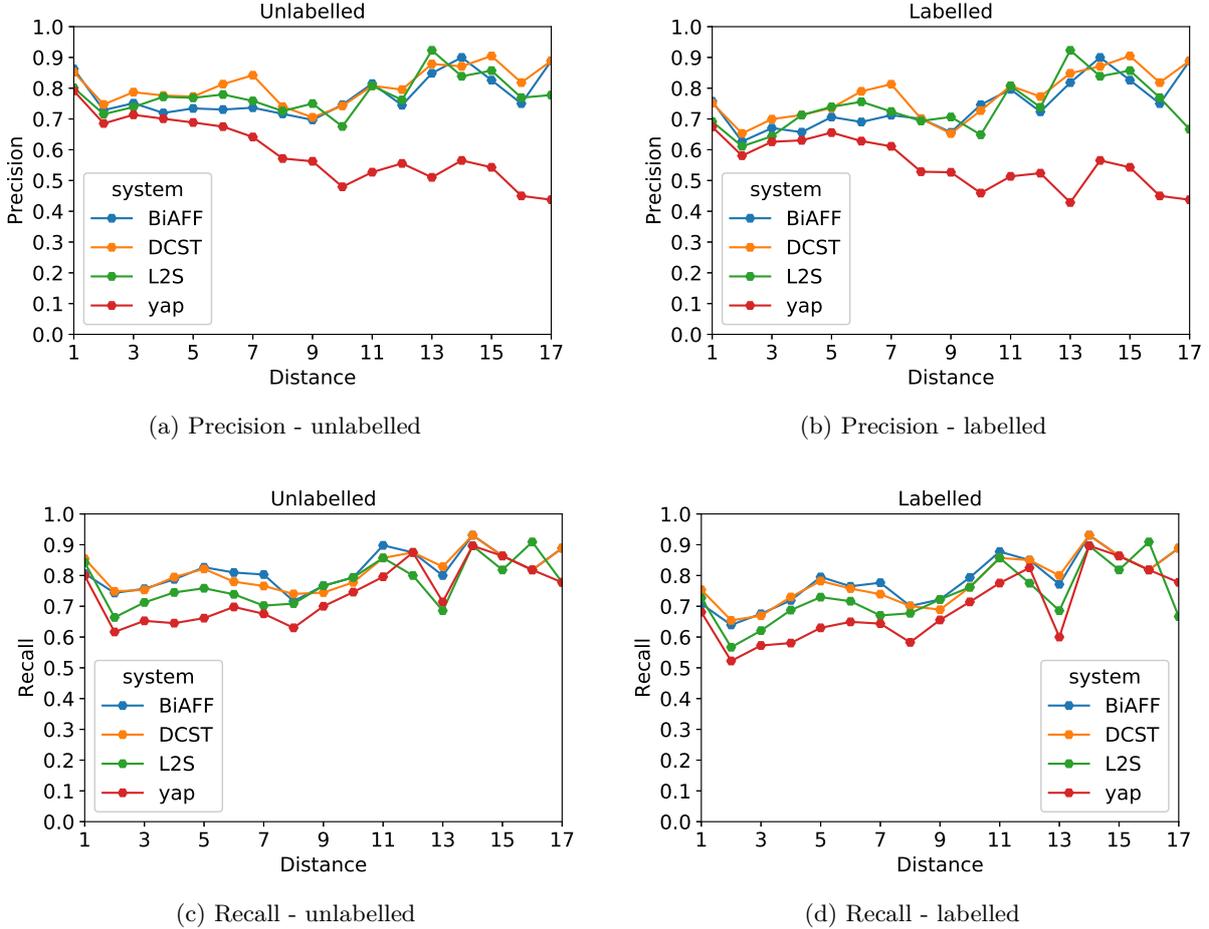

(a) Precision - unlabelled  (b) Precision - labelled

(c) Recall - unlabelled  (d) Recall - labelled

Figure 3: Precision/Recall based on unlabelled/labelled edges for STBC prose w.r.t. dependency length

**Dependency Length:** Dependency length between two related words is the distance between those words based on their linear arrangement in the sentence (McDonald and Nivre, 2007). Figure 3(a-d), shows the precision/recall for dependency length on the STBC corpus based on unlabelled/labelled edges. Transition based parsers with greedy inference, in general, tend to perform worse on longer dependencies in terms of precision (McDonald and Nivre, 2007; McDonald and Nivre, 2011), though this becomes less evident in neural transition based parsing approaches (Kulmizev et al., 2019). Both the transition based systems in our experiments, YAP and L2S, use hand-crafted features. Aligning with observations made by McDonald and Nivre (2007), YAP's performance degrades considerably when predicting longer dependencies. However, L2S performs at par with the graph based parsing approaches we experimented with. We hypothesise, this is due to the training procedure followed in L2S, where the system is not just exposed to the ground truth trajectories but also to the one step deviations to the wrong states during training. This helps mitigating the exposure bias and label bias of these systems (Wiseman and Rush, 2016).

---

[6]The mean absolute deviation of a dataset is the average distance between each data point and the mean.(Khan-Academy, accessed 15 Apr 2020)

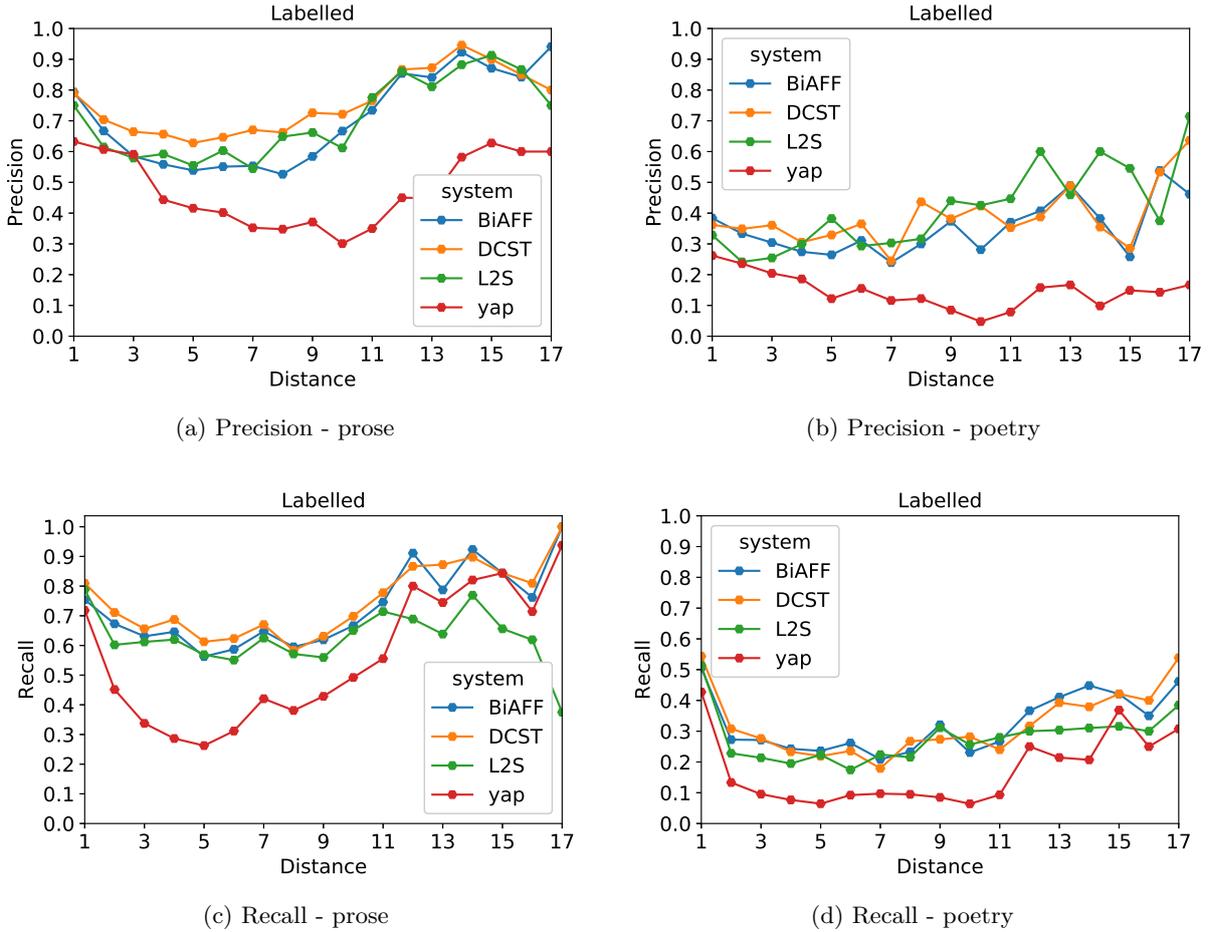

(a) Precision - prose  (b) Precision - poetry
(c) Recall - prose  (d) Recall - poetry

Figure 4: Precision/Recall based on labelled edges for Śiśupāla-vadha prose and poetry w.r.t. dependency length

Figure 4(a-d) shows the precision/recall for dependency length on the prose and poetry data on the Śiśupāla-vadha based on labelled edges. In prose order sentences, the linear arrangement of words tends to follow the principle of Sannidhi (Kulkarni et al., 2015), i.e. dependency locality (Gibson et al., 2019). In terms of precision for prose data, similar to STBC prose data, YAP under-performs consistently as compared to all other systems. In the case of precision for poetry data, we can find that DCST and L2S systems, generally provide better performance as compared to other systems on longer dependency length word pairs. From Figure 5 we can observed that STBC corpus contains about 44.9 % of the word pairs with a dependency length of one. Similarly, word pairs with a dependency length greater than five is only about 13.5 % of the total word pair connections in the ground-truth. In contrast about 30.7 % of the word pairs have a distance more than 5 in the poetry data. More importantly, these systems have less than 0.6 % word pairs with a length of 1. Given that our

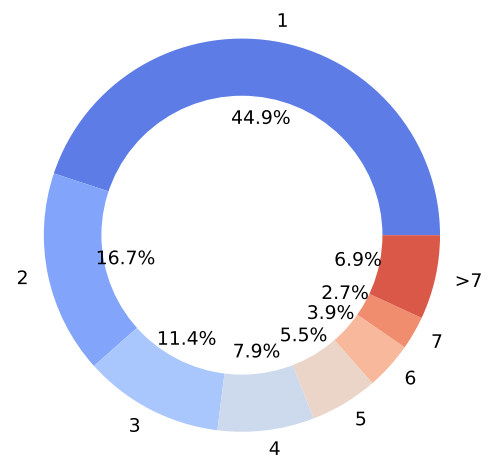

Figure 5: Dependency length percentage between word-pairs in STBC corpus ground truth

parsers are trained on prose data, these structural divergences with poetry ordered data might be a strong reason for poor performance of the parsing models on the poetry data.

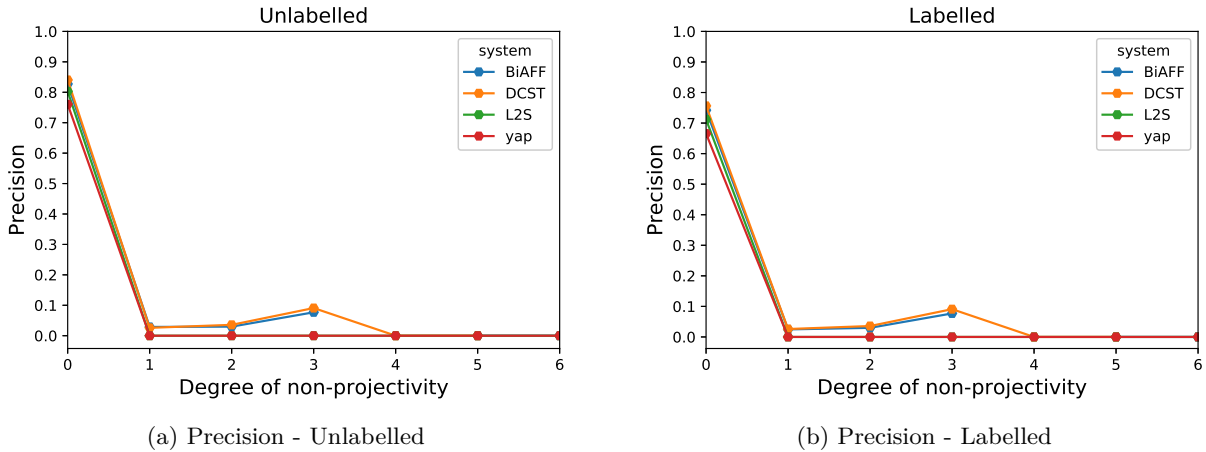

(a) Precision - Unlabelled

(b) Precision - Labelled

Figure 6: Precision based on the degree of non-projectivity of dependency arcs for STBC prose

**Non Projective Dependencies:** We further look into the performance of the systems on the non-projective dependencies. Here, we make use of degree of non-projectivity (McDonald and Nivre, 2007). The degree of non-projectivity of a dependency arc from a word $w$ to word $u$ is the number of words occurring between $w$ and $u$ which are not descendants of $w$ and modify a word not occurring between $w$ and $u$ (Nivre, 2006). Figure 6(a-b) shows the precision for different models on the STBC corpus both in terms of labelled and unlabelled predictions. Similarly, Figure 7(a-b) shows the performance of the systems on the prose and poetry ordered data from Śiśupāla-vadha. Here we only show the performance for the labelled case, and the performance for the unlabelled case is similar to that of the labelled case. It can be observed that none of the models perform well for non-projective dependency arcs (those arcs with a non-zero degree of non-projectivity). Further, for the Śiśupāla-vadha poetry data, non-projective arcs constitute 19.24 % of the edges, as compared to just 7.12 % in the corresponding prose data. This is one of the factors for low-performance of the data in poetry ordering.

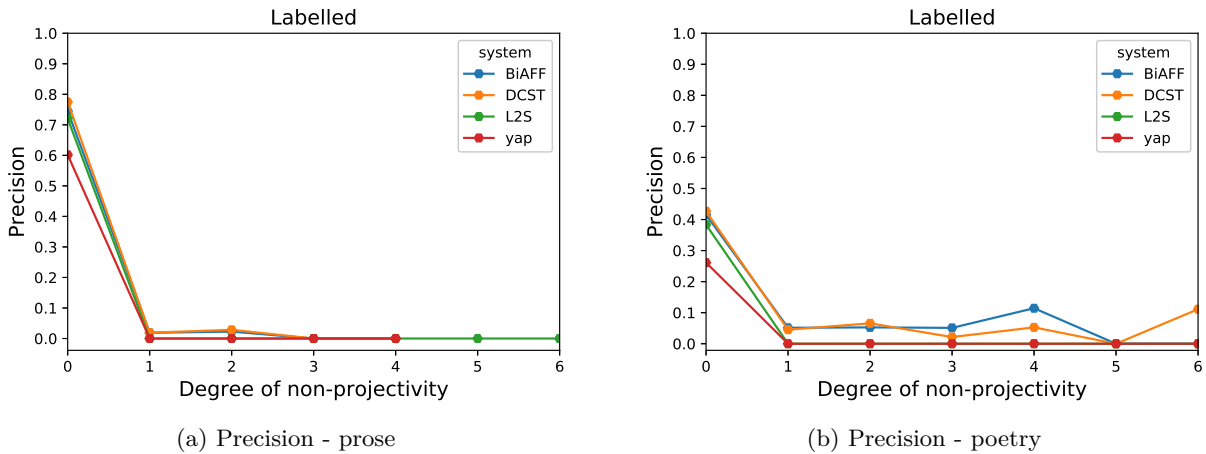

(a) Precision - prose

(b) Precision - poetry

Figure 7: Precision based on the degree of non-projectivity of dependency arcs for Śiśupāla-vadha prose and poetry

**Distance to Root:** We measure the performance of different systems based on the distance of a node from the root of the tree. Figure 8(a-d) shows the precision and recall for distance from the root on the STBC corpus based on unlabelled/labelled edges. The general trend for all the systems is that, the precision drops with nodes that have longer distances from the root. Similarly, Figure 9(a-d) shows Precision/Recall based on labelled edges for Śiśupāla-vadha prose and poetry. Transition based parsers generally tends to perform better than graph based approaches on arcs farther away from the root (McDonald and Nivre, 2011). However use of contextualised word representations tend to improve the results for graph based parsers (Kulmizev et al., 2019). Here, it needs to be observed that the graph based BiAFF parser, deteriorates in its performance with predictions that are far from the root. At the same time, DCST which uses BiAFF as its base parser reports better precision than L2S in most of the cases. We hypothesise that this is due to the improvement in the feature representations learned using the self-training approach in DCST. Further, it is interesting to note that both the graph parsers have the highest MAD on all the three dataset settings as compared to the transition based parsers. In fact, the transition based parser L2S outperforms BiAFF in most of the predictions with distance from root larger than 3, even though BiAFF reports a better overall performance than L2S.

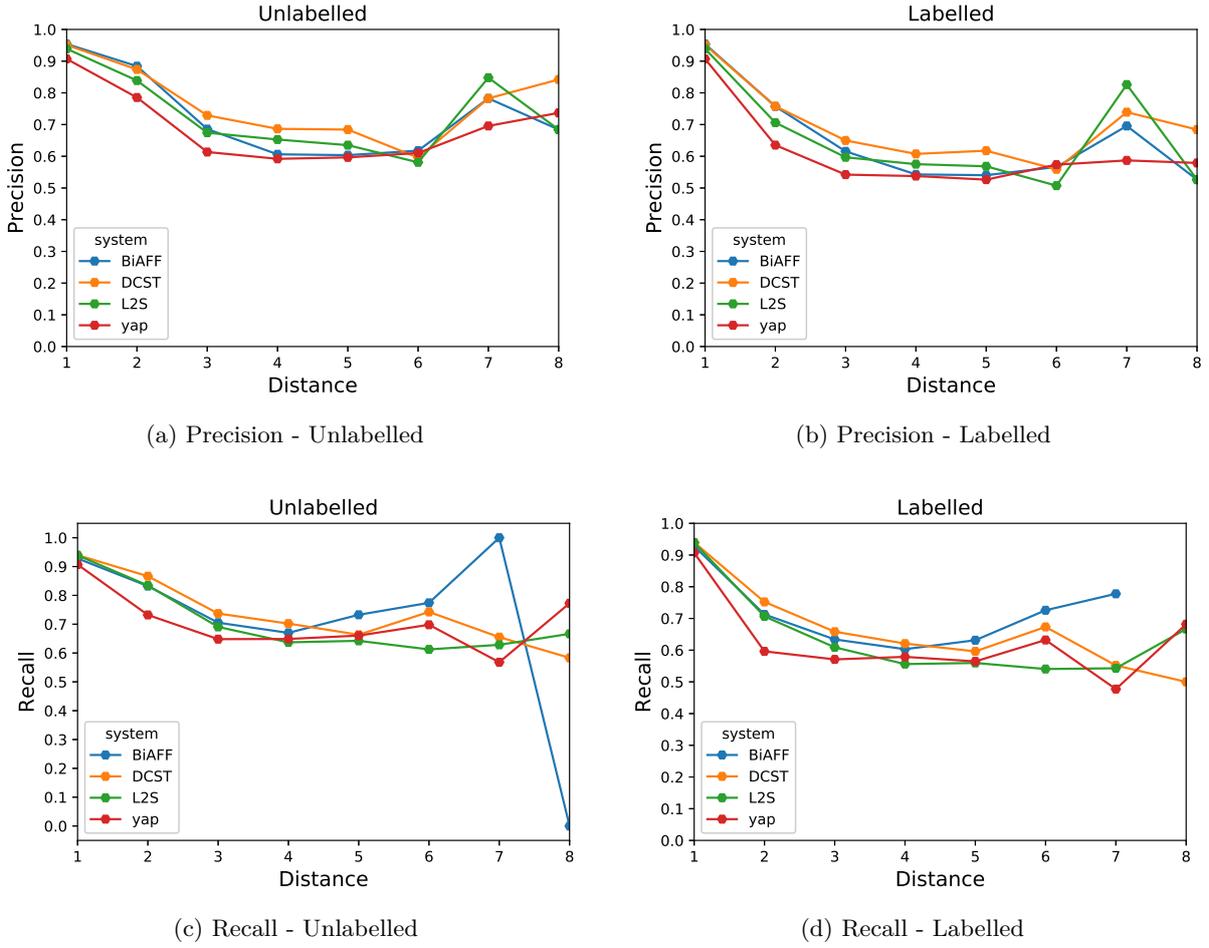

Figure 8: Precision/Recall based on unlabelled/labelled edges for STBC prose w.r.t. distance to the root

## 5 Conclusion

In this work, we evaluated four standard neural dependency parsing models for parsing of Sanskrit sentences. We find that neural parsers that make use of neural word representations and contextualised word representations can be effectively used for parsing morphologically rich languages like Sanskrit. However, word ordering in these sentences affects the performance of these systems drastically. As of now, it would be difficult to conclude, whether this is a consequence of the linguistic characteristics of the language or because of the data distribution of the training data. To elaborate further, one possibility is that the current drop in result can be due to the distributional differences in the prose training data, with which the models were trained, and the poetry data. On the other hand, it further needs to be investigated whether the performance drop has to do with something which is more linguistically inherent, and not just confining to the distributional differences. For this, we would require to train the models using dependency tagged data in verse order. It would be worthwhile to pursue if there are schemes to shuffle the word ordering in verses that would benefit in parsing these sentences. The approaches can be to completely disregard the word order and convert the input as a graph as in the case of Krishna (2019), or to use syntactic linearisers such as Krishna et al. (2019) that convert a verse to its corresponding prose order or to perform word reordering/linearisation as an auxiliary task in the training for dependency parsing (Wang et al., 2018).

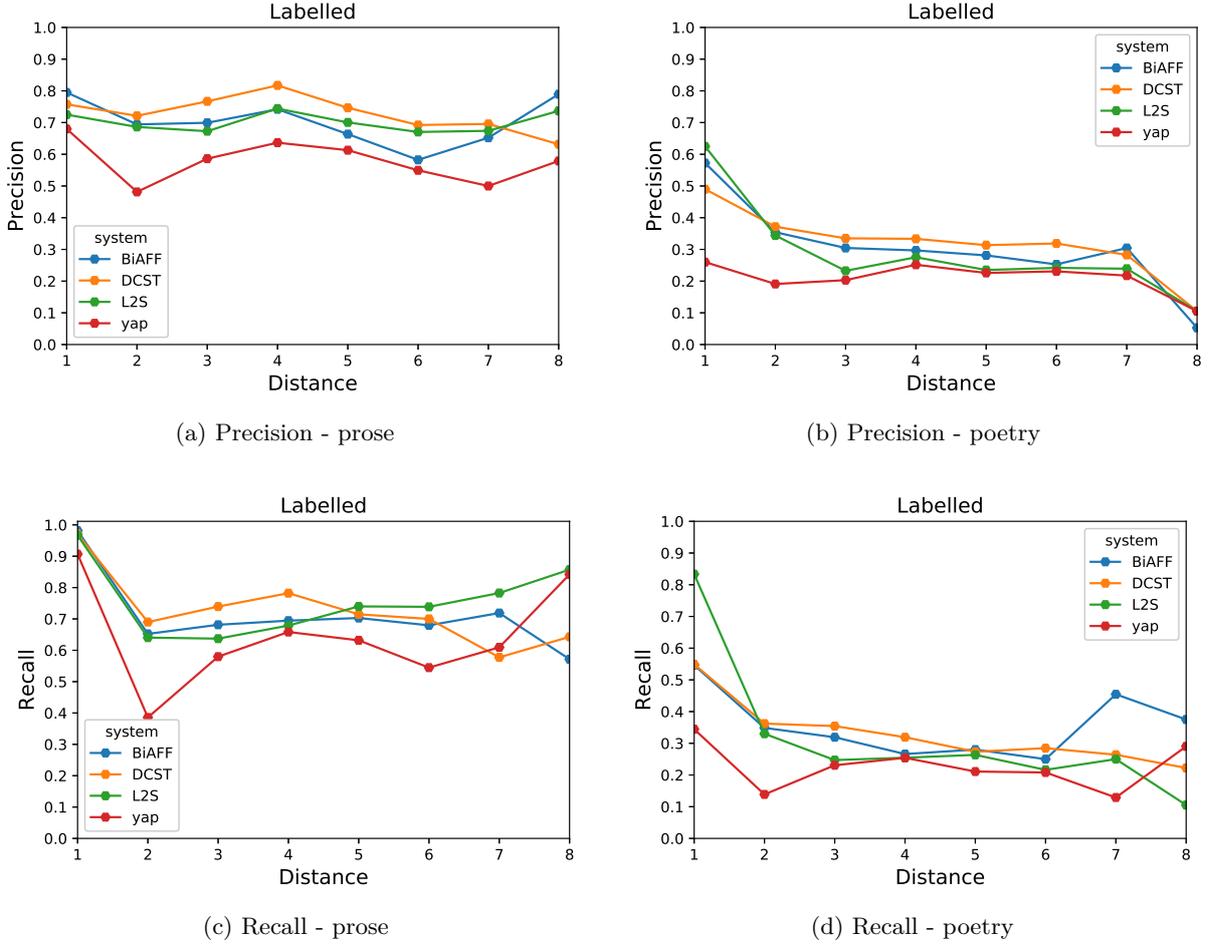

Figure 9: Precision/Recall based on labelled edges for Śiśupāla-vadha prose and poetry w.r.t. distance to the root

While in this work, we focused on evaluating the effect of standard parsers in their default settings, the next logical step would be to modify these models by incorporating design changes by keeping Sanskrit specific characteristics in mind. We plan to extend the work in two important directions. Currently, we provide the morphological tags of the words as input to the model. This often can be a bottleneck, as morphological disambiguation can be a challenge in itself for morphologically rich languages. We plan to augment the neural parsers, especially DCST, under a multi task setting where we would stack layers from a multi task morphological tagger so as to obtain enriched representations for the words. In this model, the user need not provide the morphological tags as input as the system would expect only the surface-forms of the words. Similarly, we find that the L2S model can be extended by incorporating linguistic knowledge as constraints into the inference, so as to effectively prune the search space. To elaborate, currently during one step deviation, transitions are made to incorrect states based on label predictions, including those states which can effectively be avoided by a rule based pre-processing step. The pre-processing step can bring down the search space transitions by allowing only those that are valid based on the given case information of the word pairs for which a decision needs to be taken.

## Acknowledgements

We are grateful to Anupama Ryali for sharing the Śiśupāla-vadha dataset and Amba Kulkarni for the STBC corpus.